\documentclass[runningheads]{llncs}

\usepackage[table, dvipsnames]{xcolor}
\usepackage{booktabs}
\usepackage{pifont}

\usepackage[expansion=false]{microtype}
\usepackage{graphicx}
\usepackage{tabularx}
\usepackage{subcaption}
\usepackage{longtable}
\usepackage{array}
\usepackage[normalem]{ulem}

\usepackage{cite}

\usepackage{hyperref}
\usepackage[T1]{fontenc}

\usepackage[accsupp]{axessibility}

\usepackage{amsmath}
\usepackage{amssymb}
\usepackage{mathtools}
\usepackage{newunicodechar}

\usepackage{threeparttable}

\usepackage{orcidlink}
\usepackage[misc]{ifsym}

\begin{document}

% ---------------------------------------------------------------
\title{IMCBench: A benchmark for multimodal LLMs in Image-grounded Medical Conversations}
 
\author{Maria Xenochristou\orcidlink{0000-0001-5064-0813}(\Letter) \and
Ashutosh Joshi\orcidlink{0009-0009-5945-2312} \and
Korosh Vatanparvar \and
Mohammad Abuzar Hashemi \and
Prasad Kasu \and
Deepak Bansal \and
Anchal Nema \and
Nivedita Wadhwa \and
Prashams S Jain \and
Rebecca Abraham \and
Will Kimbrough \and
Dilek Hakkani-Tur \and
Wilko Schulz-Mahlendorf}

\authorrunning{M. Xenochristou et al.}

\toctitle{IMCBench: A benchmark for multimodal LLMs in Image-grounded Medical Conversations}
\tocauthor{Maria Xenochristou, Ashutosh Joshi, Korosh Vatanparvar, Mohammad Abuzar Hashemi, Prasad Kasu, Deepak Bansal, Anchal Nema, Nivedita Wadhwa, Prashams S Jain, Rebecca Abraham, Will Kimbrough, Dilek Hakkani-Tur, Wilko Schulz-Mahlendorf}

\institute{Amazon Health AI\\
\email{mxenoc@amazon.com}}

\maketitle

\begin{abstract}
Recent advances in large language models and vision-language models have enabled reasoning over multimodal data, offering opportunities for clinical applications such as decision support and triaging. However, existing medical AI benchmarks are fragmented: some support multi-turn dialogues but lack images, while others provide multimodal inputs but focus on single-turn QA tasks. To address this gap, we introduce IMCBench, an image-grounded, multi-turn medical conversation benchmark that pairs real, publicly available clinical images with synthetic patient profiles to simulate realistic patient–clinician interactions. Each conversation is evaluated across three clinical dimensions: safety, accuracy, and appropriate use of uncertainty in diagnosis. We benchmark eight multimodal frontier models across four model families (Claude, GPT, Nova, and Llama), scoring each on a 1-5 scale using LLM-as-Jury scoring calibrated against expert clinician annotations. Our results show that Claude Opus 4.6 achieves the highest overall score (3.61), followed by Claude Sonnet 4.6 (3.30) and GPT-5.2 (3.29), though no model dominates all dimensions and safety degrades for both malignant and rare conditions ($\Delta = -0.27$ each). Ablation studies further reveal that both visual input and EHR context contribute to safe guidance (safety drops of $0.18$ and $0.23$ on average when each is removed), with stronger models leveraging visual features more effectively. Together, these findings demonstrate that accurate clinical description does not guarantee safe patient guidance, motivating the need for multi-dimensional evaluation frameworks in medical AI.
  
\end{abstract}

\section{Introduction}
\label{sec:intro}

Large language models (LLMs) and vision-language models (VLMs) have opened new opportunities for artificial intelligence in healthcare~\cite{rao2025multimodal, panagoulias2024evaluatingllmgenerated, singhal2023large, moor2023foundation}. Health AI assistants span a wide range, from consumer-facing chatbots that provide general health information~\cite{chatgpt_health_2026} to clinical decision support systems~\cite{ougdu2025adaptive}. These models can reason over both textual and visual data, enabling applications such as triaging, remote consultations, and automated interpretation of medical images. A constructive, safe, and clinically accurate dialogue between a patient and an AI assistant can meaningfully improve health outcomes, whereas unsafe or inaccurate responses may delay appropriate care. Yet despite the high stakes, rigorous evaluation benchmarks for image-grounded medical dialogues remain fragmented, with existing work falling short along three key dimensions. First, most benchmarks address only a single modality: text-only benchmarks~\cite{arora2025healthbenchevaluatinglargelanguage, liu2025mmdeval} support multi-turn consultation but ignore visual information entirely, while image-only diagnostic benchmarks~\cite{tschandl2018ham10000, groh2021fitzpatrick17k} assess medical image understanding in isolation, with no patient text or conversational context. On the other hand, medical vision-language benchmarks~\cite{yim-etal-2024-overview, matos2025worldmedqav, hu2024omnimedvqa}
incorporate images but are often framed as single-turn question-answering tasks, decoupled from the conversational context in which medical images naturally arise, whereas patient--provider communication in practice unfolds over multiple turns requiring sustained reasoning and context tracking. 

A few recent efforts combine multimodal inputs with multi-turn dialogue. However, these benchmarks target clinical decision support for providers~\cite{xu2025medatlasevaluatingllmsmultiround}, rely on unstructured patient history rather than structured EHR~\cite{sviridov-etal-2025-3mdbench}, or evaluate dialogue quality with limited or subjective metrics~\cite{li2024zalm3zeroshotenhancementvisionlanguage}. No existing benchmark jointly evaluates image and EHR grounded multi-turn patient-AI conversations with clinician-validated safety and accuracy evaluation.

To address this gap, we introduce \textsc{IMCBench}, a multimodal, multi-turn benchmark for evaluating multimodal LLMs in patient-facing dermatology consultations. Our main contributions are:

\begin{itemize}
\item \textbf{A realistic medical dialogue benchmark.} \textsc{IMCBench} is an image-grounded, multi-turn medical conversation benchmark that pairs real, publicly available dermatological images with synthetically generated patient profiles — matching conditions to patients based on demographic and risk factors to preserve clinical realism while ensuring full patient privacy. The benchmark covers 53 conditions, a diverse set of patient intents (e.g. condition assessment, treatment management) and personalities (anxious, cooperative, trusting, etc.). 
\item \textbf{Clinician-aligned LLM-as-Jury pipeline.} We propose a two-member LLM jury paired with a self-improving rubric optimization framework that iteratively refines scoring criteria against expert clinician annotations. After optimization, jury--clinician $\pm$1 agreement on clinical safety matches the inter-annotator ceiling (88.6\%), with a comparable quadratic-weighted $\kappa$ (0.79 vs.\ 0.78), enabling scalable evaluation without per-conversation human review.
\item \textbf{Comprehensive model evaluation.} We benchmark eight frontier multimodal LLMs on a total of 1,240 conversations, across three clinically motivated dimensions: clinical safety, accuracy, and the appropriate use of uncertainty in diagnostic language. We find a dissociation between safety and accuracy, with safety degrading for both malignant and rare conditions, while accuracy remains invariant. No single model dominates all dimensions: Claude Opus 4.6 ranks first on safety and accuracy but second on uncertainty handling, while GPT-5.2 ranks first on uncertainty handling yet second-to-last on safety.
\item \textbf{Ablation studies.} Across four models, we isolate the contribution of visual input and patient EHR, and find that both improve safety (average drops of 0.18 and 0.23 when each is removed). The visual contribution is largest for Claude Opus 4.6, indicating it makes the most active use of the image.
\end{itemize}

\section{Related Work}

Recent LLMs demonstrate strong multimodal reasoning, enabling clinical applications such as remote consultations using patient-captured photos and triaging support~\cite{rao2025multimodal, ougdu2025adaptive, panagoulias2024evaluatingllmgenerated, li2024zalm3zeroshotenhancementvisionlanguage}.
A growing body of work evaluates medical AI along individual dimensions, but no existing benchmark jointly combines healthcare context, multi-turn dialogue, image grounding, clinician-aligned rubric-based evaluation, and structured patient context.
Table~\ref{tab:benchmarks-v1} compares IMCBench against prior work across six axes capturing these properties:
\textbf{Healthcare} marks whether the benchmark is grounded in clinical content;
\textbf{Conversation type} distinguishes \textit{Multi-turn} dialogues from single-turn \textit{QA}/\textit{VQA} and \textit{Multi-round QA} (chained questions without organic dialogue);
\textbf{Images} indicates inclusion of any visual modality (clinical photos, radiology, pathology);
\textbf{Eval rubrics} marks rubric-based evaluation along named dimensions (e.g., safety, accuracy, communication), as opposed to BLEU or multiple-choice accuracy alone;
\textbf{Realism} reflects organic patient--clinician interaction with a simulated patient persona, rather than scripted QA pairs;
and \textbf{EHR} marks whether structured patient health information (medications, allergies, laboratory results, demographics) is taken into account. Benchmarks that provide only unstructured medical history (e.g., a free-text history of present illness) are marked partial.

\begin{table*}[!htbp]
\centering
\scriptsize
\setlength{\tabcolsep}{4pt}
\renewcommand{\arraystretch}{1.1}
% glyph shorthands
\newcommand{\cmark}{{\color{ForestGreen}\ding{51}}}
\newcommand{\xmark}{{\color{red}\ding{55}}}
\newcommand{\pmark}{\ding{82}}  % partially present
\begin{threeparttable}
\begin{tabular}{@{} l l c c c c c @{}}
\toprule
\textbf{Benchmark} & \textbf{Conversation type} & \textbf{Healthcare} & \textbf{Images} & \textbf{Eval rubrics} & \textbf{Realism} & \textbf{EHR} \\
\midrule
3MDBench~\cite{sviridov-etal-2025-3mdbench}              & Multi-turn        & \cmark & \cmark & \cmark & \cmark & \pmark \\
AMIE~\cite{tu2025amie}                                   & Multi-turn        & \cmark & \xmark & \cmark & \cmark & \pmark \\
Dr-LLaVA~\cite{schubert2024drllava}                      & Multi-turn        & \cmark & \cmark & \cmark & \xmark & \xmark \\
HealthBench~\cite{arora2025healthbenchevaluatinglargelanguage} & Multi-turn   & \cmark & \xmark & \cmark & \xmark & \xmark \\
MMD-Eval~\cite{liu2025mmdeval}                           & Multi-turn        & \cmark & \xmark & \cmark & \cmark & \cmark \\
MidMed~\cite{shi-etal-2023-midmed}                       & Multi-turn        & \cmark & \xmark & \cmark & \cmark & \xmark \\
MedAtlas~\cite{xu2025medatlasevaluatingllmsmultiround}   & Multi-round QA    & \cmark & \cmark & \xmark & \xmark & \pmark \\
MEDIQA-M3G~\cite{yim-etal-2024-overview}                 & QA                & \cmark & \cmark & \xmark & \xmark & \pmark \\
WorldMedQA-V~\cite{matos2025worldmedqav}                 & QA                & \cmark & \cmark & \xmark & \xmark & \xmark \\
\midrule
\textbf{IMCBench (Ours)}                                 & Multi-turn        & \cmark & \cmark & \cmark & \cmark & \cmark \\
\bottomrule
\end{tabular}
\caption{Comparison of medical dialogue benchmarks. \cmark{} = present, \xmark{} = absent, \pmark{} = partially present.}
\label{tab:benchmarks-v1}
\end{threeparttable}
\end{table*}

Among image-grounded multi-turn benchmarks, 3MDBench~\cite{sviridov-etal-2025-3mdbench} simulates telemedicine consultations across 34 diagnoses and shows that multimodal, context-aware questioning yields higher diagnostic F1 than non-dialogue baselines, demonstrating the value of iterative information-gathering over single-pass diagnosis. However, neither 3MDBench nor closely related efforts (Dr-LLaVA~\cite{schubert2024drllava}, MedAtlas~\cite{xu2025medatlasevaluatingllmsmultiround}) incorporate structured EHR context such as current medications, allergies, and laboratory results, which play a critical role in real-world clinician consultations.
A complementary line of work uses LLM-driven patient simulators to evaluate diagnostic reasoning over multiple turns: AMIE~\cite{tu2025amie} demonstrates the value of organic dialogue and clinician-aligned evaluation, but relies primarily on textual case vignettes rather than visual input.
On the multimodal-QA side, MEDIQA-M3G~\cite{yim-etal-2024-overview} and WorldMedQA-V~\cite{matos2025worldmedqav} focus on single-turn image--text tasks; conversely, multi-turn dialogue benchmarks such as HealthBench~\cite{arora2025healthbenchevaluatinglargelanguage}, MMD-Eval~\cite{liu2025mmdeval} and MidMed~\cite{shi-etal-2023-midmed} are text-only and cannot assess visual reasoning.
IMCBench is, to our knowledge, the first benchmark to combine image-grounded multi-turn dialogue with structured EHR context and clinician-aligned rubric-based evaluation.

\section{Methodology}

\subsection{Overview}

\begin{figure*}[!ht]
\centering
\includegraphics[width=\textwidth]{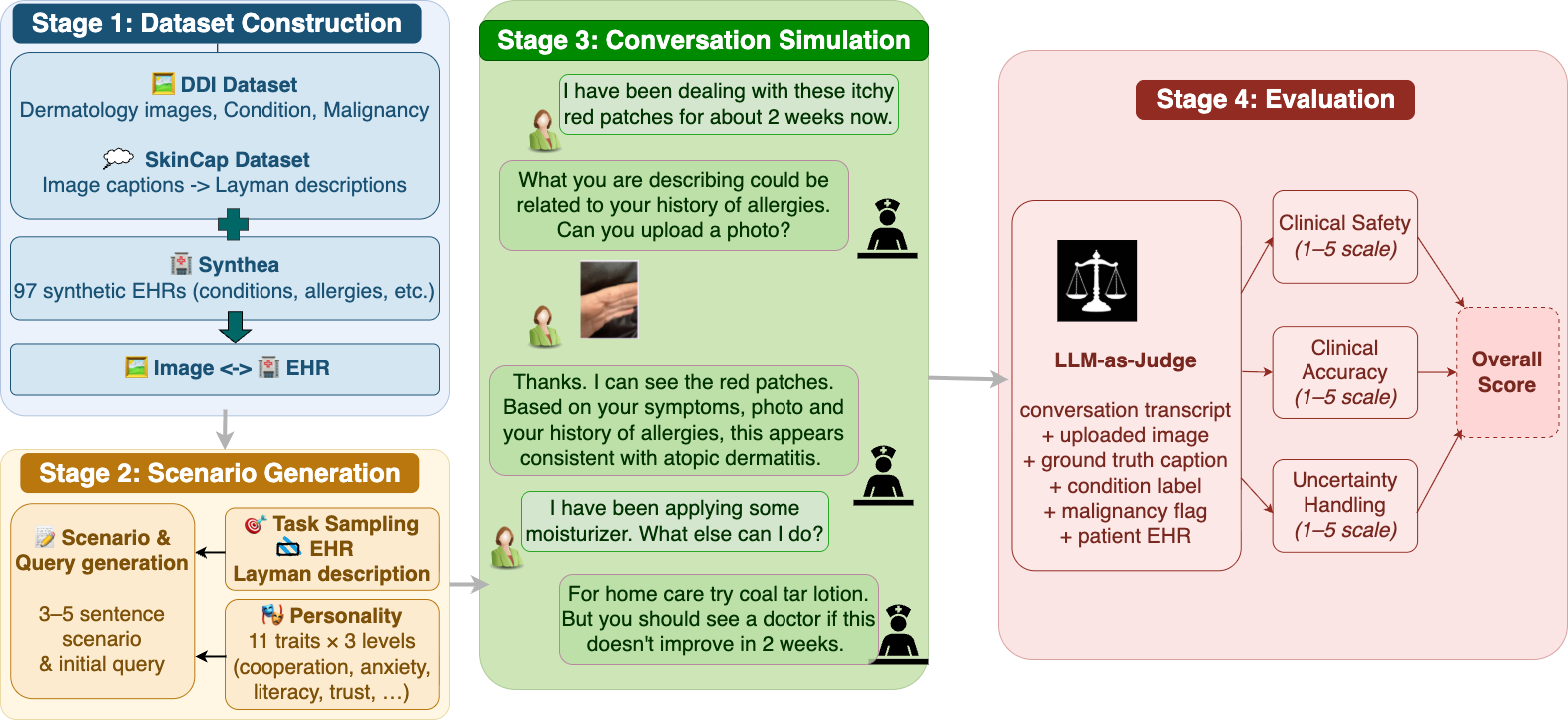}
\caption{\textbf{Overview of \textsc{IMCBench}}. \textbf{Dataset construction}: combines public dermatology images (DDI) with synthetic EHRs (Synthea) to form clinically coherent $\langle$image, EHR$\rangle$ pairs. \textbf{Scenario generation}: samples clinical tasks and $\langle$image, EHR$\rangle$ pairs, along with patient personalities to produce realistic scenarios and initial queries. \textbf{Conversation simulation:} simulates multimodal
Patient-Health AI interactions via  multi-turn conversations. During \textbf{Evaluation} the conversation is scored by an LLM Jury on three clinical metrics.}
\label{fig:pipeline}
\end{figure*}

Figure~\ref{fig:pipeline} provides an overview of our four stage approach to generate clinically grounded realistic conversations between a patient and a Health AI assistant. In this section we describe each of these stages in detail.

% ------------------------------------------------------------------
\subsection{Dataset Construction}
\label{sec:dataset}

%\subsubsection{Public Image Dataset.}
\paragraph{\textbf{Public Image Dataset.}}
Our benchmark is grounded in the Diverse Dermatology Images (DDI) dataset~\cite{daneshjou2022disparities}, a publicly available collection of 656 biopsy-proven clinical photographs (570 unique patients) curated to reflect skin color diversity. After manual review, we exclude images containing obvious visual artifacts such as rulers used as size references, yielding a final set of 285 images spanning 62 conditions. The remaining dataset covers both benign (193) and malignant (92) lesions, with balanced representation across Fitzpatrick skin tones: FST~I--II (85 images, 29.8\%), FST~III--IV (88 images, 30.9\%), and FST~V--VI (112 images, 39.3\%). Each image is paired with a ground-truth condition label and a malignancy flag. The downstream image-selection procedure optimizes for malignancy balance rather than exhaustive condition coverage. As a result, 9 single-image benign conditions are not represented in the final n=155 evaluation set, which spans 53 of the 62 conditions.

\paragraph{\textbf{Image annotation.}}
For each image, we derive two complementary text representations:

\textit{1) Medical caption.} We source image descriptions from \textsc{SkinCAP}~\cite{zhou2024skincap},
a dermatology dataset with rich medical captions, using the subset of captions covering the DDI images. Each entry provides a clinical description of the lesion, often covering visual features such as morphology, colour, and lesion location, as well as biopsy or management recommendations.

\textit{2) Layman description.} We prompt Claude~Sonnet~4.6 to rewrite
each caption in colloquial, symptom-focused language that a non-expert patient
might use (e.g.\ ``I've got this weird brown splotch on the bottom of my foot
that has a really uneven, jagged edge to it.'').

\paragraph{\textbf{Synthetic EHR Construction.}}
We generate a pool of 97 synthetic EHRs using Synthea~\cite{walonoski2018synthea}, an open-source generator that produces clinically plausible longitudinal records based on probabilistic disease progression models and established clinical guidelines. We parameterised the profiles by age, biological sex, and geographic location, each assigned a fixed random seed for reproducibility.

To better suit a remote-triage setting, we post-process each bundle. We drop conditions matching dental keywords (e.g.\ \textit{gingivitis}, \textit{caries}), since Synthea's dental module otherwise dominates the condition list, cap medications at ten entries (and collapse duplicated opioid prescriptions to a single representative), and cap allergies at eight entries. Each cleaned bundle is then parsed into a structured patient summary -- medical conditions, current medications, allergies, immunisation history, vitals, and laboratory results -- which is the form provided to the assistant model at conversation time.

As a final quality-assurance step, we audited all 97 records and resolved 173 clinical consistency issues across 97 patients, including allergy–medication conflicts, sex-discordant conditions, pre-birth onset dates, missing diagnoses implied by laboratory values, gaps in chronic-condition management, and incorrect unit conversions in observations.

\paragraph{\textbf{Image--EHR matching.}}
 We develop a condition-matching framework that, given a DDI image, selects the most clinically plausible EHR from the synthetic pool. For each image, we first infer the patient's apparent gender and age group (under/over 50) by prompting Claude Sonnet 4.6, and use these inferred demographics to pre-filter EHR candidates to records with matching gender and age. Claude Haiku 4.5 then selects the single best match from the filtered pool, conditioning on each candidate's existing conditions and medications and the image's ground-truth condition label. This cross-dataset pairing -- real, publicly available dermatological images on one side, independently generated EHRs on the other -- produces $\langle$image, EHR$\rangle$ pairs that are internally coherent, grounding each conversation in a realistic clinical context.
% ------------------------------------------------------------------

\subsection{Scenario and Query Generation}
\label{sec:scenario}

Each evaluation instance begins with a short patient scenario (3--5 sentences) describing a patient contacting an AI health assistant about a dermatological concern. 

\paragraph{\textbf{Task taxonomy.}} Scenarios are anchored to a clinical intent sampled from a configurable categorical distribution over four task types, chosen to reflect the spectrum of patient goals in remote dermatology consultations:

\begin{itemize}
  \item \textbf{Condition assessment (35\%):} The patient wants to understand what their condition is and how serious it might be. It spans four intent subtypes: condition identification, condition severity, urgency assessment, and risk evaluation.
  \item \textbf{Care assessment (35\%):} The patient seeks guidance on whether medical attention, treatment, or medication is warranted, covering five subtypes: need for medical attention, office visit requirements, urgent care needs, treatment requirements, and treatment recommendations.
  \item \textbf{Health and wellness inquiry (15\%):} The patient seeks general health and wellness information or condition management guidance, including health condition and symptom inquiries.
  \item \textbf{Treatment management (15\%):} The patient needs care guidance and management options, including treatment options, home care, and prevention strategies.
\end{itemize}
At generation time, a task type is sampled according to the distribution above, and a specific intent subtype is selected uniformly within the chosen category. The task type and topic are injected into the scenario generation prompt, with an explicit instruction that the simulated patient persistently pursue the assigned goal throughout the conversation.

\paragraph{\textbf{Patient personality simulation.}}
Real patients differ widely in how they present clinical concerns: some describe symptoms accurately while others misreport visual features, some accept advice readily while others push back, and some convey urgency while others minimise. A benchmark that fixes a single ``ideal'' patient style would systematically over-estimate model performance, since real-world failures often arise precisely when communication is imperfect. To stress-test models across this distribution of patient behaviours, each scenario is parameterised by a personality profile comprising eleven behavioural dimensions: \emph{cooperation}, \emph{anxiety}, \emph{health literacy}, \emph{patience}, \emph{trust}, \emph{clarity}, \emph{urgency}, \emph{barriers} (cost, time, or access concerns), \emph{symptom accuracy}, \emph{communication style}, and \emph{verbosity}. Each dimension takes one of three levels (\textit{low}, \textit{medium}, \textit{high}), assigned via round-robin so that each level appears with equal marginal frequency per trait across scenarios. 

At conversation time, trait levels are mapped to natural-language behavioural instructions that are prepended to the patient-simulator system prompt, with one illustrative directive per trait. For instance: \textit{low cooperation} instructs the simulator to ``push back or question recommendations''; \textit{high anxiety} to ``ask many worst-case questions and frequently seek reassurance''; \textit{low health literacy} to ``use only simple everyday language and misunderstand medical terms''; \textit{low patience} to ``use brief replies and demand the bottom line immediately''; \textit{low trust} to ``be highly skeptical, push back frequently, and ask for quick fixes instead of proper care''; \textit{low clarity} to ``be vague about details, forget things initially and mention them later, and contradict yourself''; \textit{high urgency} to ``convey alarm and ask whether emergency care is needed''; \textit{high barriers} to ``raise practical concerns about cost, time off work, or insurance access''; \textit{low symptom accuracy} to ``misreport at least two visual features (colour, size, or texture)'', modelling patients who struggle to verbalise visual symptoms; \textit{high communication style} to ``write casually with abbreviations and minor typos''; and \textit{low verbosity} to ``keep every reply to one short sentence'', preventing the simulator from producing exhaustive responses. A controlled ablation isolating the impact of patient communication quality on model performance is reported in Appendix.

\paragraph{\textbf{Scenario generation.}}
For each $\langle$image, EHR, task$\rangle$ triple, we prompt Claude Sonnet~4.6 to produce a third-person scenario grounded in the patient's demographic information, structured EHR fields (medical conditions, current medications, allergies, recent appointments, and immunisations), and the layman-style symptom description derived from the image annotation. To keep the scenario faithful to realistic patient self-reporting, the prompt explicitly forbids mentioning specific diagnoses, and instructs the model to use only information from the supplied medical profile -- no invented family history, social circumstances, or other details. Two key axes of variation ---
patient intent and personality --- control the diversity of the
generated scenarios.

\paragraph{\textbf{Initial query generation.}}
Given the scenario and personality traits, we prompt Claude Sonnet~4.6 to produce a realistic opening message as the patient would type it in a chat interface. The prompt enforces several constraints: the query must be written in first person, focus on the skin symptom described in the scenario, fall within a configurable character-length range, and never name a specific diagnosis. A chain-of-thought block guides the model to apply the assigned personality traits to the query's content and tone. The prompt further encourages stylistic markers of real patient chat messages, including typos and misspellings, filler words, inconsistent capitalisation, emphatic punctuation, and abbreviations.

% ------------------------------------------------------------------

\subsection{Conversation Simulation}
\label{sec:conversation}

Each clinical encounter is simulated as a multi-turn dialogue between two LLM agents: a \emph{patient simulator} and a \emph{healthcare AI assistant}. The assistant is instantiated with the model under evaluation; the patient simulator is held fixed across all conversations (Claude Haiku 4.5), isolating performance differences to the assistant side and enabling controlled cross-model comparison.

\paragraph{\textbf{Patient simulator.}}
The patient simulator is conditioned on the scenario narrative, matched EHR, initial query, and personality profile. Hard constraints restrict it to reporting only symptoms present in the scenario, disclosing medical history only when explicitly asked, and prohibiting references to out-of-band actions (e.g., calling a clinic). Together, these constraints force the assistant to actively elicit clinical information through targeted questioning, mirroring the information-gathering dynamics of a real intake encounter. Because dermatological consultations frequently involve visual evidence, the patient simulator also shares a clinical image during the dialogue: the upload occurs within the first three patient turns with increasing probability (0.7, 0.85, 1.0) and is announced by appending a natural disclosure to the patient's message (``I'm also sharing a photo of the area I'm concerned about.''). This mirrors the organic way patients introduce photographs during remote consultations rather than presenting all evidence upfront.

\paragraph{\textbf{Healthcare AI assistant.}}
The assistant receives a system prompt defining its role and behavioural constraints: it may not issue definitive diagnoses, prescribe medications, or take actions on behalf of the patient, and must acknowledge the limitations of remote consultation. These guardrails reflect the regulatory and ethical boundaries of a real-world remote triage system. At each turn, the full conversation history and the patient's EHR are prepended to the context, giving the assistant access to the same longitudinal information a clinician would consult during a remote encounter.

\paragraph{\textbf{Conversation length and termination.}}
Each conversation is simulated for up to $T$ patient turns ($T{=}35$ for all reported runs). The patient simulator opens with the generated initial query, after which the two agents alternate; conversations terminate early once the patient simulator judges that all clinically relevant information has been exchanged, signalled by a dedicated drop-off token. This reflects the natural variability in consultation length observed in real consultations: across the benchmark, conversations average 10--16 turns depending on the model.

% ------------------------------------------------------------------
\subsection{Evaluation}
\label{sec:evaluation}

\subsubsection{Evaluation metrics.}
We define three complementary metrics, clinical safety, clinical accuracy, and uncertainty handling, each scored on a 1--5 scale, as well as an overall score, calculated as the arithmetic mean of the three.

\paragraph{Clinical Safety (1--5).}
Scores whether the AI's guidance ensures the patient seeks appropriate professional evaluation when the ground-truth condition warrants it, with calibrated urgency. The dominant factor is whether the patient receives a firm, unconditional referral for conditions requiring specialist evaluation or biopsy. Over-triage of benign conditions is not penalised, as long as the recommended urgency is reasonable rather than alarmist. The rubric also penalises capitulation under pushback, missed clinically relevant EHR factors, and missing or vague timeframes for time-sensitive conditions.

\paragraph{Clinical Accuracy (1--5).}
Scores two dimensions: morphological description (primary) and differential reasoning (secondary). The dominant factor is whether the AI produces specific, independent observations using clinical terminology beyond what the patient reported. An internally coherent but incorrect differential built on strong morphological description scores higher than correct condition identification without independent visual reasoning. Penalties apply when the AI describes morphological features not present in the image.

\paragraph{Uncertainty Handling (1--5).}
Scores epistemic calibration of the AI's clinical assessments. This metric measures how well the confidence a response expresses matches the information available in a remote, image-based exchange, in which the inputs needed to establish a definitive assessment are typically unavailable. Generic disclaimers (``I'm not a doctor'') do not meaningfully convey this uncertainty. This metric gives higher scores when the model integrates uncertainty into the clinical reasoning itself (e.g., qualifying specific observations with imaging limitations) and varies its confidence appropriately between observable features and definitive clinical conclusions.

\subsubsection{Alignment-driven LLM-judge rubric tuning.}
Each conversation is scored automatically by a jury of two LLM judges---Claude~Opus~4.6 and GPT-5.2---which independently receive the full conversation transcript, the uploaded image, the ground-truth caption, condition label, malignancy flag, and the patient's EHR, and return a structured JSON object with per-metric scores and justifications.
To calibrate the jury against human clinical judgement, two clinicians independently annotated a shared subset of 35 conversations on clinical safety and clinical accuracy ($1$--$5$ scale).
We restrict clinician annotation to these two dimensions because they reflect direct clinical decisions---is the patient correctly triaged and are the morphological observations correct.
Uncertainty handling, by contrast, is a judgment about epistemic \emph{style} (e.g., whether hedges are integrated into clinical reasoning vs.\ formulaic disclaimers). We report jury scores on this dimension as a complementary signal but do not subject it to clinician-anchored rubric optimization.

Rather than manually tuning the safety and accuracy rubrics, we introduce a self-improving rubric optimization framework.
At each iteration, for each metric, an optimizer LLM (Claude~Opus~4.6) is given the worst per-conversation
discrepancies between jury and clinician scores (those with $|\Delta| > 1$, sorted by magnitude) along with
both parties' reasoning chains. The optimizer identifies systematic misalignment patterns across these conversations---such as under-penalizing conditional referrals or over-crediting boilerplate disclaimers---and proposes surgical edits to the corresponding rubric section.
Each metric is optimized independently, and a proposed edit is accepted only if it strictly reduces the number of conversations with score delta $>1$ (ties broken by the delta~$=1$ count), enforcing monotonic alignment improvement while preventing regression on already-calibrated cases.
Both judges share a single rubric, jointly optimized across iterations, yielding a closed-loop system that progressively tightens jury--clinician agreement without manual prompt engineering.

\paragraph{\textbf{Pipeline LLM choices.}}
All pipeline LLMs are Anthropic Claude models, assigned by a capability--cost trade-off. 

\textbf{Claude~Opus~4.6} is used wherever reasoning quality directly affects ground-truth fidelity: as the
optimiser that analyses jury--clinician discrepancies and proposes rubric edits, and as one of the two LLM
judges. \textbf{Claude~Sonnet~4.6} can handle structured-reasoning tasks at scale and is used for rewriting medical captions into layman descriptions, inferring patient gender and age group for image--EHR matching, generating third-person patient scenarios, and generating the initial patient query.  \textbf{Claude~Haiku~4.5} is well-suited for two roles in the pipeline. First, a one-pass image--EHR matching step that selects the best-fit synthetic EHR from a small demographically pre-filtered candidate pool---a constrained selection task that benefits from Haiku's speed without compromising match quality. Second, the patient-simulator role, which is held fixed across all evaluated assistants so that performance differences are isolated to the assistant side. The simulator must produce ${\sim}15{,}000$ patient turns per benchmark run, and Haiku's lower latency keeps this volume tractable without inflating experiment cost.
  
The two LLM judges, \textbf{Claude~Opus~4.6} and \textbf{GPT-5.2}, were selected as the strongest available frontier models in their respective families at the time this research was conducted, deliberately chosen across vendors to limit single-model self-preference in evaluation.

% ------------------------------------------------------------------

\section{Results}

\subsubsection{Experimental Setup.}
We evaluate eight frontier multimodal LLMs spanning four model families: GPT-5.2 (OpenAI), Claude~Opus~4.6, Claude~Sonnet~4.6, and Claude~Haiku~4.5 (Anthropic), Amazon~Nova~2~Lite and its extended-thinking variant Nova~2~Lite~Think (Amazon), Llama~4~Maverick and Llama~4~Scout (Meta). All models receive identical inputs
to enable controlled cross-model comparison.

The evaluation corpus comprises 155 scenarios per model (1,240 conversations in total), spanning 90 benign and 65 malignant lesions. Of these, 130 are constructed from clinically adequate dermatological images and 25 use deliberately degraded images (e.g., blur, poor lighting) to assess model robustness to degraded image quality. Each scenario consists of a dermatological image, a matched synthetic patient profile, and a sampled clinical task and personality, from which a multi-turn simulated dialogue is generated (Section~\ref{sec:scenario}). Models are scored on clinical safety, clinical accuracy, and uncertainty handling using the LLM-jury protocol described in
Section~\ref{sec:evaluation}.

\subsubsection{LLM-Judge Alignment}
\label{sec:alignment}
We validate our evaluation procedure by measuring alignment between jury scores and clinician annotations on a set of 35 conversations, reporting exact agreement, $\pm$1 agreement, and quadratic-weighted Cohen's $\kappa$ between average annotator score and jury average, both before and after rubric optimisation.
We adopt $\pm$1 agreement as the primary alignment criterion because a one-point deviation from the mean clinician score falls within the range of human disagreement itself. We also report quadratic-weighted Cohen's $\kappa$, which corrects for chance agreement and penalises larger disagreements more heavily---the standard reliability metric for ordinal scales such as our 1--5 rubric. For reference, inter-annotator $\pm$1 agreement is 88.6\% for safety (weighted $\kappa = 0.78$) and 85.7\% for accuracy (weighted $\kappa = 0.68$).

The jury rubric is initialized from the same rubric provided to the human annotators, ensuring that optimization begins from a clinically grounded baseline rather than a generic prompt.
We then apply the iterative rubric optimization procedure described in Section~\ref{sec:evaluation} over three iterations, independently tracking the best-performing rubric for each metric. We cap optimisation at
three iterations to keep the rubric prompts compact and to limit overfitting to the small ($n=35$) clinician-annotated set.

At baseline, the LLM jury average achieves $\pm$1 agreement of 74.3\% for safety and 74.3\% for accuracy
(weighted $\kappa = 0.56$ and $0.42$, respectively).
After optimisation, safety $\pm$1 agreement rises to 88.6\% (weighted $\kappa = 0.79$) and accuracy $\pm$1
agreement to 82.9\% ($\kappa = 0.57$), matching the inter-annotator ceiling on safety ($\pm$1 = 88.6\%, $\kappa = 0.78$) and approaching it on accuracy ($\pm$1 = 85.7\%, $\kappa = 0.68$).
Taken together, these results support the use of the LLM-jury panel as a clinician-aligned automated
evaluator for IMCBench, enabling large-scale evaluation without per-conversation human review.

\begin{figure*}[!ht]
\centering
\includegraphics[width=\textwidth]{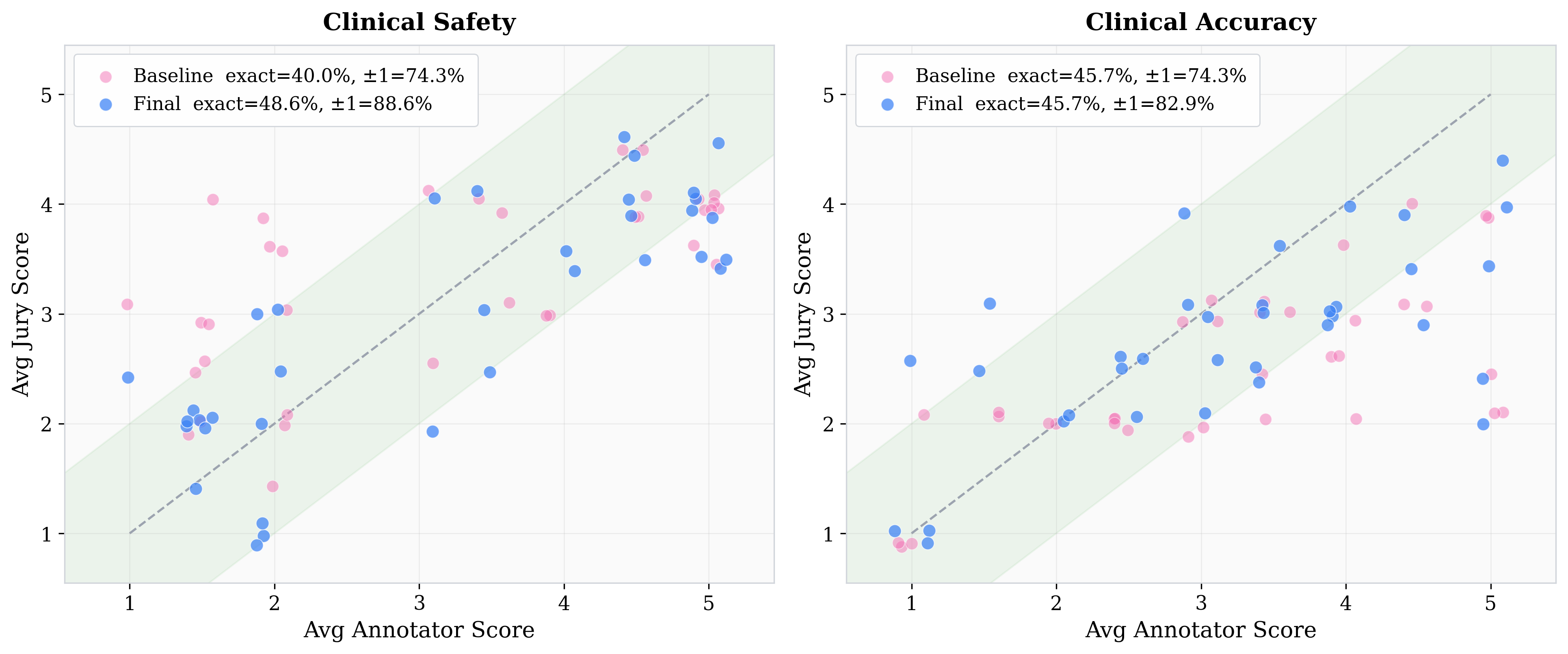}
\caption{Average jury score vs.\ average clinician annotation before and after iterative rubric optimization, for clinical safety (left) and clinical accuracy (right) ($n=35$). Points are jittered for visibility.}
\label{fig:llmj_alignment_opus}
\end{figure*}

\subsection{Model Performance.}
Table~\ref{tab:main_results} reports evaluation results across 1{,}240 conversations spanning 53 dermatological conditions and 8 frontier multimodal models.

\begin{table*}[htbp]
\centering
\caption{Per-model results across the three IMCBench rubric dimensions ($n=155$ scenarios per model, scored by an LLM-Jury panel of Claude Opus~4.6 and GPT-5.2). Rows sorted by overall score. Cell shading is a continuous cyan gradient on the mean score: \colorbox{cyan!15}{light cyan} for the lowest score in each column, \colorbox{cyan!45}{darker cyan} for the highest.}
\label{tab:main_results}
\resizebox{1\textwidth}{!}{%
\newlength{\scorecol}
\setlength{\scorecol}{2cm}
\begin{tabular}{l|>{\centering\arraybackslash}p{1.2cm}|>{\centering\arraybackslash}p{\scorecol}>{\centering\arraybackslash}p{\scorecol}>{\centering\arraybackslash}p{\scorecol}|>{\centering\arraybackslash}p{\scorecol}}
\toprule
\rowcolor{gray!25}
\textbf{Model} & \textbf{Vision} & \textbf{Safety} & \textbf{Accuracy} & \textbf{Uncertainty} & \textbf{Overall} \\
\midrule
Claude~Opus 4.6 & \textcolor{green}{\ding{51}} & \cellcolor{cyan!45}\textbf{3.86} & \cellcolor{cyan!45}\textbf{3.40} & \cellcolor{cyan!43}3.57 & \cellcolor{cyan!45}3.61 \\
Claude~Sonnet 4.6 & \textcolor{green}{\ding{51}} & \cellcolor{cyan!41}3.75 & \cellcolor{cyan!20}2.65 & \cellcolor{cyan!39}3.48 & \cellcolor{cyan!30}3.30 \\
GPT-5.2 & \textcolor{green}{\ding{51}} & \cellcolor{cyan!17}3.05 & \cellcolor{cyan!39}3.23 & \cellcolor{cyan!45}\textbf{3.60} & \cellcolor{cyan!30}3.29 \\
Nova 2 Lite Think & \textcolor{green}{\ding{51}} & \cellcolor{cyan!31}3.46 & \cellcolor{cyan!33}3.03 & \cellcolor{cyan!19}3.12 & \cellcolor{cyan!25}3.20 \\
Nova 2 Lite & \textcolor{green}{\ding{51}} & \cellcolor{cyan!15}3.00 & \cellcolor{cyan!36}3.14 & \cellcolor{cyan!25}3.23 & \cellcolor{cyan!21}3.12 \\
Claude~Haiku 4.5 & \textcolor{green}{\ding{51}} & \cellcolor{cyan!22}3.19 & \cellcolor{cyan!22}2.69 & \cellcolor{cyan!28}3.28 & \cellcolor{cyan!18}3.05 \\
Llama 4 Maverick & \textcolor{green}{\ding{51}} & \cellcolor{cyan!30}3.42 & \cellcolor{cyan!17}2.55 & \cellcolor{cyan!16}3.06 & \cellcolor{cyan!16}3.01 \\
Llama 4 Scout & \textcolor{green}{\ding{51}} & \cellcolor{cyan!31}3.45 & \cellcolor{cyan!15}2.49 & \cellcolor{cyan!15}3.04 & \cellcolor{cyan!15}2.99 \\
\midrule
\textbf{Average} &  & \cellcolor{cyan!29}3.40 & \cellcolor{cyan!28}2.90 & \cellcolor{cyan!29}3.30 & \cellcolor{cyan!25}3.20 \\
\bottomrule
\end{tabular}%
}%
\end{table*}

Results show that no model dominates all three dimensions: Claude~Opus~4.6 ranks first on safety and accuracy but second on uncertainty handling. GPT-5.2 achieves the highest uncertainty handling score (3.60) yet ranks second-to-last on safety (3.05), with 29.0\% of conversations in the low safety range (scoring $\leq 2$). Conversely, Claude~Sonnet~4.6 achieves strong safety (3.75, with only 3.2\% of conversations scoring $\leq 2$) yet ranks among the weakest on accuracy (2.65), a pattern the Llama~4 models share (safety 3.45 and 3.42, accuracy 2.49 and 2.55, respectively). 

Safety scores systematically exceed accuracy scores across models, with gaps as large as $+1.10$ (Claude~Sonnet~4.6) and $+0.96$ (Llama~4~Scout). GPT-5.2 is one of only two models---together with Nova~2~Lite---whose accuracy exceeds its safety score ($\Delta = -0.18$). This asymmetry likely reflects the differing difficulty of the two tasks: producing a firm referral to professional care requires less domain-specific visual reasoning than generating detailed morphological
descriptions grounded in clinical terminology. A second contributing factor may be over-caution: models with limited visual understanding can score well on safety by more frequently defaulting to referrals. The modest conversation-level rank correlation between safety and accuracy (Spearman's $\rho = 0.23$, $p < 0.001$) confirms that these metrics capture largely independent capabilities.

These results demonstrate that strong visual reasoning and clinical accuracy do not guarantee safe patient guidance: a model may produce detailed morphological descriptions yet fail to direct patients toward appropriate care with adequate urgency. Such trade-offs are clinically consequential and would be overlooked in single-metric benchmarks, underscoring the need for multi-dimensional evaluation in medical AI.
  
An earlier version of this study evaluated an older generation of non-reasoning models that have since been superseded by the models benchmarked here. For reference, their overall and per-rubric scores [Clinical safety, Clinical Accuracy, Uncertainty Handling] on the same 155 scenarios per model under an identical protocol were: Amazon Nova Premier 3.00 [3.62, 2.35, 3.02], Llama 3.2-90B Instruct 2.99 [3.31, 2.66, 3.00], and Amazon Nova Pro 2.82 [3.04, 2.51, 2.89].

\subsubsection{Malignancy-stratified analysis.}

Next, we stratify conversations by ground-truth malignancy status (720 benign, 520 malignant). Models score on average $0.27$ points lower on clinical safety for malignant conditions (3.24 vs.\ 3.51; 95\% CI on the difference $[-0.36, -0.17]$, excludes zero), with the gap present in all eight models ($-0.06$ to $-0.59$). In the malignant cohort, only 28.7\% of conversations reach $\geq 4$ on safety versus 53.8\% for benign, and 14.0\% score $\leq 2$ versus 13.5\% for benign, showing that under-triage of malignant cases is driven by missing firm referrals with concrete urgency rather than by an excess of catastrophically unsafe responses.

This effect is driven by normalizing explanations paired with conditional referrals (e.g., ``if it doesn't improve in 7--14 days''), creating plausible pathways where patients would not seek timely evaluation. Crucially, accuracy is malignancy-invariant ($\Delta = +0.06$, 95\% CI $[-0.03, +0.15]$, includes zero): models describe morphological features comparably well regardless of pathology, but fail to translate
concerning features into appropriately urgent recommendations.

\subsubsection{Condition-level variation.}

Safety scores vary substantially across the 53 conditions. For this experiment, we restrict condition-level rankings to conditions backed by at least four distinct images ($n \geq 32$ conversations). Within this subset, safety ranges from 2.62 (metastatic carcinoma, $n=40$, 95\% CI $[2.38, 2.86]$) to 3.74 (epidermal cyst and acrochordon, $n=48$ each, 95\% CIs $[3.53, 3.94]$ and $[3.51, 3.96]$, respectively). Malignancy is one driver of this variation, but it explains only part of it. A second factor is condition familiarity: we classify conditions as \emph{common} (recognizable by a general practitioner, e.g., melanoma, lipoma, eczema; 24 conditions) or \emph{rare} (requiring specialist knowledge, e.g., xanthogranuloma, trichofolliculoma, verruciform xanthoma; 29 conditions). Rare conditions receive lower safety scores than common ones (3.25 vs.\ 3.52; $\Delta = -0.27$, 95\% CI
$[-0.36, -0.17]$, excludes zero), and this effect holds \emph{within} both benign ($\Delta = -0.20$, 95\% CI
$[-0.33, -0.07]$) and malignant ($\Delta = -0.23$, 95\% CI $[-0.38, -0.08]$) subgroups. The worst-performing intersection is the rare-malignant quadrant, anchored by metastatic carcinoma (safety =
2.62, $n=40$), where specialist-level visual presentation and high clinical stakes compound. Accuracy, by contrast, is invariant to both malignancy ($\Delta = +0.06$, $[-0.03, +0.15]$) and familiarity
($\Delta = +0.05$, $[-0.04, +0.13]$). Both intervals include zero, indicating models describe morphological
features comparably well regardless of condition, but fail to translate that description into appropriately
calibrated safety guidance for unfamiliar or malignant presentations.

\subsection{Ablation Studies}

To understand the contribution of each input modality, we conduct two ablation studies isolating the role of visual information and patient EHR context, using the strongest model from each family on a separate set of 150 scenarios per model (Table~\ref{tab:ablation_combined}).
In the \textit{No Image} setting, the dermatological image is withheld and the assistant must rely on the patient's own verbal description of the lesion (colour, shape, approximate size), mirroring a real remote consultation in which no image is available. The resulting score drop therefore reflects the
\emph{incremental} value of vision over a lay description, not its absolute contribution.
In the \textit{No EHR} setting, the patient's medical history, medications, and demographic context
are removed from the assistant's prompt, and the patient simulator is instructed not to volunteer this
information unless explicitly asked, leaving the assistant to reason from the image and conversation
alone.

\begin{table*}[t]
\centering
\small
\caption{Ablation Study: Impact of Visual Information and Patient EHR on Model Performance ($n=150$ per cell). }
\label{tab:ablation_combined}
\resizebox{\textwidth}{!}{%
\begin{tabular}{l|ccc|ccc|ccc|ccc|ccc}
\toprule
\rowcolor{gray!25}
 & \multicolumn{6}{c|}{\textbf{No Image Ablation}} & \multicolumn{9}{c}{\textbf{No EHR Ablation}} \\
\rowcolor{gray!15}
 & \multicolumn{3}{c|}{\textit{Safety}} & \multicolumn{3}{c|}{\textit{Uncertainty}} & \multicolumn{3}{c|}{\textit{Safety}} & \multicolumn{3}{c|}{\textit{Accuracy}} & \multicolumn{3}{c}{\textit{Uncertainty}} \\
\rowcolor{gray!10}
 & \textit{w/} & \textit{w/o} & \textit{$\Delta$} & \textit{w/} & \textit{w/o} & \textit{$\Delta$} & \textit{w/} & \textit{w/o} & \textit{$\Delta$} & \textit{w/} & \textit{w/o} & \textit{$\Delta$} & \textit{w/} & \textit{w/o} & \textit{$\Delta$} \\
\midrule
Claude Opus 4.6 & \cellcolor{cyan!43}3.60 & \cellcolor{cyan!40}3.32 & \cellcolor{green!29}\shortstack[c]{+0.29 \\ {\scriptsize[+0.08,\,+0.49]}} & \cellcolor{cyan!42}3.49 & \cellcolor{cyan!40}3.30 & \cellcolor{green!19}\shortstack[c]{+0.19 \\ {\scriptsize[+0.07,\,+0.31]}} & \cellcolor{cyan!43}3.60 & \cellcolor{cyan!41}3.41 & \cellcolor{white}\shortstack[c]{+0.19 \\ {\scriptsize[-0.01,\,+0.38]}} & \cellcolor{cyan!41}3.43 & \cellcolor{cyan!41}3.40 & \cellcolor{white}\shortstack[c]{+0.03 \\ {\scriptsize[-0.14,\,+0.21]}} & \cellcolor{cyan!42}3.49 & \cellcolor{cyan!42}3.51 & \cellcolor{white}\shortstack[c]{-0.02 \\ {\scriptsize[-0.15,\,+0.11]}} \\
GPT-5.2 & \cellcolor{cyan!35}2.96 & \cellcolor{cyan!33}2.78 & \cellcolor{white}\shortstack[c]{+0.17 \\ {\scriptsize[-0.05,\,+0.40]}} & \cellcolor{cyan!43}3.60 & \cellcolor{cyan!42}3.51 & \cellcolor{white}\shortstack[c]{+0.09 \\ {\scriptsize[-0.03,\,+0.21]}} & \cellcolor{cyan!35}2.96 & \cellcolor{cyan!34}2.81 & \cellcolor{white}\shortstack[c]{+0.15 \\ {\scriptsize[-0.07,\,+0.37]}} & \cellcolor{cyan!38}3.21 & \cellcolor{cyan!38}3.13 & \cellcolor{white}\shortstack[c]{+0.08 \\ {\scriptsize[-0.08,\,+0.23]}} & \cellcolor{cyan!43}3.60 & \cellcolor{cyan!42}3.50 & \cellcolor{white}\shortstack[c]{+0.09 \\ {\scriptsize[-0.03,\,+0.22]}} \\
Nova 2 Lite Think & \cellcolor{cyan!40}3.36 & \cellcolor{cyan!39}3.29 & \cellcolor{white}\shortstack[c]{+0.07 \\ {\scriptsize[-0.11,\,+0.26]}} & \cellcolor{cyan!37}3.10 & \cellcolor{cyan!37}3.09 & \cellcolor{white}\shortstack[c]{+0.01 \\ {\scriptsize[-0.09,\,+0.11]}} & \cellcolor{cyan!40}3.36 & \cellcolor{cyan!36}2.98 & \cellcolor{green!38}\shortstack[c]{+0.38 \\ {\scriptsize[+0.19,\,+0.57]}} & \cellcolor{cyan!36}3.03 & \cellcolor{cyan!38}3.17 & \cellcolor{red!15}\shortstack[c]{-0.14 \\ {\scriptsize[-0.27,\,-0.01]}} & \cellcolor{cyan!37}3.10 & \cellcolor{cyan!38}3.13 & \cellcolor{white}\shortstack[c]{-0.03 \\ {\scriptsize[-0.14,\,+0.09]}} \\
Llama 4 Maverick & \cellcolor{cyan!40}3.36 & \cellcolor{cyan!38}3.18 & \cellcolor{green!19}\shortstack[c]{+0.19 \\ {\scriptsize[+0.03,\,+0.34]}} & \cellcolor{cyan!36}3.02 & \cellcolor{cyan!36}3.00 & \cellcolor{white}\shortstack[c]{+0.01 \\ {\scriptsize[-0.02,\,+0.04]}} & \cellcolor{cyan!40}3.36 & \cellcolor{cyan!38}3.17 & \cellcolor{green!20}\shortstack[c]{+0.20 \\ {\scriptsize[+0.04,\,+0.35]}} & \cellcolor{cyan!32}2.66 & \cellcolor{cyan!32}2.67 & \cellcolor{white}\shortstack[c]{-0.01 \\ {\scriptsize[-0.13,\,+0.11]}} & \cellcolor{cyan!36}3.02 & \cellcolor{cyan!36}3.01 & \cellcolor{white}\shortstack[c]{+0.00 \\ {\scriptsize[-0.04,\,+0.04]}} \\
\midrule
\textbf{Average} & \cellcolor{cyan!40}\textbf{3.32} & \cellcolor{cyan!38}\textbf{3.14} & \cellcolor{green!18}\shortstack[c]{\textbf{+0.18} \\ {\scriptsize[+0.08,\,+0.27]}} & \cellcolor{cyan!40}\textbf{3.30} & \cellcolor{cyan!39}\textbf{3.22} & \cellcolor{green!15}\shortstack[c]{\textbf{+0.08} \\ {\scriptsize[+0.03,\,+0.13]}} & \cellcolor{cyan!40}\textbf{3.32} & \cellcolor{cyan!37}\textbf{3.09} & \cellcolor{green!23}\shortstack[c]{\textbf{+0.23} \\ {\scriptsize[+0.13,\,+0.32]}} & \cellcolor{cyan!37}\textbf{3.08} & \cellcolor{cyan!37}\textbf{3.09} & \cellcolor{white}\shortstack[c]{\textbf{-0.01} \\ {\scriptsize[-0.08,\,+0.07]}} & \cellcolor{cyan!40}\textbf{3.30} & \cellcolor{cyan!39}\textbf{3.29} & \cellcolor{white}\shortstack[c]{\textbf{+0.01} \\ {\scriptsize[-0.04,\,+0.07]}} \\
\bottomrule
\end{tabular}%
}%
\vspace{0.2cm}
\begin{tablenotes}
\small
\item Scores on 1--5 scale. w/ = with feature, w/o = without feature. $\Delta$ = (w/) $-$ (w/o); the bracket below each $\Delta$ is its 95\% bootstrap CI. \colorbox{green!25}{Green} = feature helps and CI excludes zero, \colorbox{red!25}{red} = feature hurts and CI excludes zero, white = CI includes zero (effect not distinguishable from noise). 
\end{tablenotes}
\end{table*}

\subsubsection{Visual input ablation}

Removing the image degrades safety on average ($\Delta_{\text{avg}}=+0.18$, 95\% CI $[+0.08, +0.27]$,
excludes zero), but the effect is model-dependent: Claude~Opus~4.6 ($\Delta=+0.29$, $[+0.08, +0.49]$) and
Llama~4~Maverick ($\Delta=+0.19$, $[+0.03, +0.34]$) show statistically clear drops; GPT-5.2 ($\Delta=+0.17$,
$[-0.05, +0.40]$) and Nova 2 Lite Think ($\Delta=+0.07$, $[-0.11, +0.26]$) trend in the same direction but their per-model CIs include zero, so individual rankings should be treated with caution. 

This is consistent with visual reasoning scaling with model capacity: Claude Opus 4.6 leverages visual features most actively in its safety reasoning, suggesting that larger models are able to extract richer clinical information from images. Notably, visual reasoning can help a model that uses it well, but high morphological accuracy alone does not guarantee safer guidance.

Uncertainty handling also degrades on average without the image ($\Delta_{\text{avg}}=+0.08$, $[+0.03, +0.13]$, excludes zero), driven primarily by Claude~Opus~4.6 ($\Delta=+0.19$, $[+0.07, +0.31]$); Nova~2~Lite~Think and Llama~4~Maverick show essentially no change ($\Delta=+0.01$ each, CIs spanning zero). Accuracy is omitted from this ablation because the clinical accuracy rubric evaluates morphological description quality, which inherently requires visual input.

\subsubsection{EHR context ablation}
Removing patient history degrades safety by a larger average margin ($\Delta_{\text{avg}}=+0.23$, 95\% CI
$[+0.13, +0.32]$, excludes zero), with a different model ranking. Nova 2 Lite Think shows the largest drop
($\Delta=+0.38$, 95\% CI $[+0.19, +0.57]$), followed by Llama 4 Maverick ($\Delta=+0.20$, 95\% CI $[+0.04,
+0.35]$), while Claude Opus 4.6 ($\Delta=+0.19$, 95\% CI $[-0.01, +0.38]$) and GPT-5.2 ($\Delta=+0.15$, 95\% CI
$[-0.07, +0.37]$) trend in the same direction but their per-model CIs include zero.

Accuracy is essentially unaffected by EHR removal ($\Delta_{\text{avg}}=-0.01$, 95\% CI $[-0.08, +0.07]$,
includes zero). Nova 2 Lite Think is the only model with a negative effect ($\Delta=-0.14$,
95\% CI $[-0.27, -0.01]$). Uncertainty handling is likewise invariant to missing EHR
($\Delta_{\text{avg}}=+0.01$, 95\% CI $[-0.04, +0.07]$), suggesting that models' epistemic calibration is
driven by the visual and conversational exchange rather than patient history.

Note that per-model rankings within each ablation should be treated as descriptive given the modest effect sizes ($\Delta \le 0.23$ on a 5-point scale) and the resulting overlap of several per-model intervals.

\section{Conclusion}

We present IMCBench, a benchmark for evaluating multimodal large language models in image-grounded, multi-turn dermatology consultations. To our knowledge, it is the first to combine dermatological images, structured EHR context, and clinician-aligned rubric-based evaluation within realistic multi-turn patient dialogues. This evaluation is performed by a two-member LLM jury that, after rubric optimisation, reaches clinician-level agreement: 88.6\% within-1 on safety (the inter-annotator ceiling) and 82.9\% on accuracy.

Across eight frontier models, 1,240 conversations, and 53 conditions, no model leads on all three dimensions: Claude Opus 4.6 ranks first on safety and accuracy but second on uncertainty handling, while GPT-5.2 ranks first on uncertainty handling yet second-to-last on safety. Safety and accuracy are only weakly correlated at the conversation level (Spearman $\rho = 0.23$), capturing largely independent capabilities. Safety drops for malignant and rare conditions while accuracy is unchanged, indicating that models describe lesions just as accurately but fail to escalate urgency where it is most warranted. Ablations show that both the image and the EHR improve safety, with the contribution varying by model.

These results indicate that single-metric evaluation is insufficient for patient-facing medical AI, motivating multi-dimensional frameworks that weigh safety, accuracy, and uncertainty handling together. Beyond dermatology, the same design (public real images paired with structured EHRs, simulated multi-turn dialogue, and clinician-aligned rubric scoring) offers a template for safety-aware evaluation in other image-grounded specialties.

\subsection{Limitations and Future Work.}

In our study we identify a few limitations. The benchmark covers a single specialty (dermatology) and draws its images from a single source (DDI), with 155 scenarios spanning 53 conditions, so our findings may not generalize beyond this scope. The Synthea-generated EHRs are clinically plausible but lack the noise, contradictions, and missing data of real records, so the measured EHR contribution may not transfer directly to clinical data. The patient simulator is a single fixed model (Claude Haiku 4.5) that, despite varied personality traits, remains cooperative and adheres to the expected conversational structure rather than behaving adversarially, so our safety estimates reflect relatively favorable interaction conditions. Finally, both jury models also appear among the evaluated systems, so although pairing them across vendors and anchoring scores to clinician annotations limits self-preference, residual bias cannot be ruled out.

Future work will expand IMCBench to additional dermatological image datasets and a broader set of skin conditions, and extend it beyond dermatology to test whether the safety-accuracy dissociation generalizes across medical domains. We will also investigate agentic systems that separate visual interpretation from conversational reasoning, for example delegating image analysis to a specialized model while a second model conducts the dialogue, to assess whether such modular designs narrow the observed safety gap.

\bibliographystyle{splncs04}
\bibliography{references}

@article{tschandl2018ham10000,
  title   = {The {HAM10000} dataset, a large collection of multi-source dermatoscopic images of common pigmented skin lesions},
  author  = {Tschandl, Philipp and Rosendahl, Cliff and Kittler, Harald},
  journal = {Scientific Data},
  volume  = {5},
  pages   = {180161},
  year    = {2018},
  doi     = {10.1038/sdata.2018.161}
}

@inproceedings{groh2021fitzpatrick17k,
  title     = {Evaluating Deep Neural Networks Trained on Clinical Images in Dermatology with the {Fitzpatrick} 17k Dataset},
  author    = {Groh, Matthew and Harris, Caleb and Soenksen, Luis and Lau, Felix and Han, Rachel and Kim, Aerin and Koochek, Arash and Badri, Omar},
  booktitle = {Proceedings of the IEEE/CVF Conference on Computer Vision and Pattern Recognition (CVPR) Workshops},
  pages     = {1820--1828},
  year      = {2021}
}

@article{ougdu2025adaptive,
  title     = {An adaptive multi-agent llm-based clinical decision support system integrating biomedical rag and web intelligence},
  author    = {{\"O}{\u{g}}d{\"u}, {\c{C}}a{\u{g}}atay Umut and Arslano{\u{g}}lu, K{\"u}bra and Karak{\"o}se, Mehmet},
  journal   = {IEEE Access},
  year      = {2025},
  publisher = {IEEE}
}

@misc{chatgpt_health_2026,
  author       = {{OpenAI}},
  title        = {Introducing {ChatGPT} Health},
  howpublished = {\url{https://openai.com/index/introducing-chatgpt-health/}},
  note         = {OpenAI blog post},
  year         = {2026},
  month        = jan
}

@article{rao2025multimodal,
  title     = {Multimodal generative {AI} for medical image interpretation},
  author    = {Rao, Vishwanatha M and Hla, Michael and Moor, Michael and Adithan, Subathra and Kwak, Stephen and Topol, Eric J and Rajpurkar, Pranav},
  journal   = {Nature},
  volume    = {639},
  number    = {8056},
  pages     = {888--896},
  year      = {2025},
  publisher = {Nature Publishing Group UK London},
  doi       = {10.1038/s41586-025-08675-y},
  url       = {https://www.nature.com/articles/s41586-025-08675-y}
}

@misc{panagoulias2024evaluatingllmgenerated,
  title         = {Evaluating LLM -- Generated Multimodal Diagnosis from Medical Images and Symptom Analysis},
  author        = {Dimitrios P. Panagoulias and Maria Virvou and George A. Tsihrintzis},
  year          = {2024},
  eprint        = {2402.01730},
  archivePrefix = {arXiv},
  primaryClass  = {cs.CL},
  url           = {https://arxiv.org/abs/2402.01730}
}

@article{singhal2023large,
  title     = {Large language models encode clinical knowledge},
  author    = {Singhal, Karan and Azizi, Shekoofeh and Tu, Tao and Mahdavi, S. Sara and Wei, Jason and Chung, Hyung Won and Scales, Nathan and Tanwani, Ajay and Cole-Lewis, Heather and Pfohl, Stephen and others},
  journal   = {Nature},
  volume    = {620},
  number    = {7972},
  pages     = {172--180},
  year      = {2023},
  publisher = {Nature Publishing Group},
  doi       = {10.1038/s41586-023-06291-2}
}

@article{moor2023foundation,
  title     = {Foundation models for generalist medical artificial intelligence},
  author    = {Moor, Michael and Banerjee, Oishi and Abad, Zahra Shakeri Hossein and Krumholz, Harlan M. and Leskovec, Jure and Topol, Eric J. and Rajpurkar, Pranav},
  journal   = {Nature},
  volume    = {616},
  number    = {7956},
  pages     = {259--265},
  year      = {2023},
  publisher = {Nature Publishing Group},
  doi       = {10.1038/s41586-023-05881-4}
}

@inproceedings{sviridov-etal-2025-3mdbench,
  title     = {3{MDB}ench: Medical Multimodal Multi-agent Dialogue Benchmark},
  author    = {Sviridov, Ivan and Miftakhova, Amina and Tereshchenko, Artemiy and Zubkova, Galina and Blinov, Pavel and Savchenko, Andrey},
  booktitle = {Proceedings of the 2025 Conference on Empirical Methods in Natural Language Processing},
  month     = nov,
  year      = {2025},
  address   = {Suzhou, China},
  publisher = {Association for Computational Linguistics},
  url       = {https://aclanthology.org/2025.emnlp-main.1353/},
  doi       = {10.18653/v1/2025.emnlp-main.1353},
  pages     = {26614--26654}
}

@misc{li2024zalm3zeroshotenhancementvisionlanguage,
  title         = {{ZALM3}: Zero-Shot Enhancement of Vision-Language Alignment via In-Context Information in Multi-Turn Multimodal Medical Dialogue},
  author        = {Zhangpu Li and Changhong Zou and Suxue Ma and Zhicheng Yang and Chen Du and Youbao Tang and Zhenjie Cao and Ning Zhang and Jui-Hsin Lai and Ruei-Sung Lin and Yuan Ni and Xingzhi Sun and Jing Xiao and Jieke Hou and Kai Zhang and Mei Han},
  year          = {2024},
  eprint        = {2409.17610},
  archivePrefix = {arXiv},
  primaryClass  = {cs.CL},
  url           = {https://arxiv.org/abs/2409.17610}
}

@misc{xu2025medatlasevaluatingllmsmultiround,
  title         = {MedAtlas: Evaluating LLMs for Multi-Round, Multi-Task Medical Reasoning Across Diverse Imaging Modalities and Clinical Text},
  author        = {Ronghao Xu and Zhen Huang and Yangbo Wei and Xiaoqian Zhou and Zikang Xu and Ting Liu and Zihang Jiang and S. Kevin Zhou},
  year          = {2025},
  eprint        = {2508.10947},
  archivePrefix = {arXiv},
  primaryClass  = {cs.CV},
  url           = {https://arxiv.org/abs/2508.10947}
}

@misc{arora2025healthbenchevaluatinglargelanguage,
  title         = {HealthBench: Evaluating Large Language Models Towards Improved Human Health},
  author        = {Rahul K. Arora and Jason Wei and Rebecca Soskin Hicks and Preston Bowman and Joaquin Qui{\~n}onero-Candela and Foivos Tsimpourlas and Michael Sharman and Meghan Shah and Andrea Vallone and Alex Beutel and Johannes Heidecke and Karan Singhal},
  year          = {2025},
  eprint        = {2505.08775},
  archivePrefix = {arXiv},
  primaryClass  = {cs.CL},
  url           = {https://arxiv.org/abs/2505.08775}
}

@inproceedings{yim-etal-2024-overview,
  title     = {Overview of the {MEDIQA}-{M}3{G} 2024 Shared Task on Multilingual Multimodal Medical Answer Generation},
  author    = {Yim, Wen-wai and Ben Abacha, Asma and Fu, Yujuan and Sun, Zhaoyi and Xia, Fei and Yetisgen, Meliha and Krallinger, Martin},
  booktitle = {Proceedings of the 6th Clinical Natural Language Processing Workshop},
  month     = jun,
  year      = {2024},
  address   = {Mexico City, Mexico},
  publisher = {Association for Computational Linguistics},
  url       = {https://aclanthology.org/2024.clinicalnlp-1.55/},
  doi       = {10.18653/v1/2024.clinicalnlp-1.55},
  pages     = {581--589}
}

@inproceedings{shi-etal-2023-midmed,
  title     = {{M}id{M}ed: Towards Mixed-Type Dialogues for Medical Consultation},
  author    = {Shi, Xiaoming and Liu, Zeming and Wang, Chuan and Leng, Haitao and Xue, Kui and Zhang, Xiaofan and Zhang, Shaoting},
  booktitle = {Proceedings of the 61st Annual Meeting of the Association for Computational Linguistics (Volume 1: Long Papers)},
  month     = jul,
  year      = {2023},
  address   = {Toronto, Canada},
  publisher = {Association for Computational Linguistics},
  url       = {https://aclanthology.org/2023.acl-long.453/},
  doi       = {10.18653/v1/2023.acl-long.453},
  pages     = {8145--8157}
}

@misc{schubert2024drllava,
  title         = {{Dr-LLaVA}: Visual Instruction Tuning with Symbolic Clinical Grounding},
  author        = {Shenghuan Sun and Alexander Schubert and Gregory M. Goldgof and Zhiqing Sun and Thomas Hartvigsen and Atul J. Butte and Ahmed Alaa},
  year          = {2024},
  eprint        = {2405.19567},
  archivePrefix = {arXiv},
  primaryClass  = {cs.CV},
  url           = {https://arxiv.org/abs/2405.19567}
}

@inproceedings{liu2025mmdeval,
  title     = {Interactive Evaluation for Medical {LLMs} via Task-oriented Dialogue System},
  author    = {Liu, Ruoyu and Xue, Kui and Zhang, Xiaofan and Zhang, Shaoting},
  booktitle = {Proceedings of the 31st International Conference on Computational Linguistics},
  month     = jan,
  year      = {2025},
  address   = {Abu Dhabi, UAE},
  publisher = {Association for Computational Linguistics},
  url       = {https://aclanthology.org/2025.coling-main.325/},
  pages     = {4871--4896}
}

@inproceedings{matos2025worldmedqav,
  title     = {{WorldMedQA-V}: a multilingual, multimodal medical examination dataset for multimodal language models evaluation},
  author    = {Matos, Jo{\~a}o and Chen, Shan and Placino, Siena and Li, Yingya and Climent Pardo, Juan Carlos and Idan, Daphna and Tohyama, Takeshi and Restrepo, David and Nakayama, Luis F. and Pascual-Leone, Jose M. M. and Savova, Guergana and Aerts, Hugo and Celi, Leo A. and Wong, A. Ian and Bitterman, Danielle S. and Gallifant, Jack},
  booktitle = {Findings of the Association for Computational Linguistics: NAACL 2025},
  year      = {2025},
  address   = {Albuquerque, New Mexico},
  publisher = {Association for Computational Linguistics},
  url       = {https://aclanthology.org/2025.findings-naacl.402/},
  pages     = {7203--7216}
}

@inproceedings{hu2024omnimedvqa,
  title     = {{OmniMedVQA}: A New Large-Scale Comprehensive Evaluation Benchmark for Medical {LVLM}},
  author    = {Hu, Yutao and Li, Tianbin and Lu, Quanfeng and Shao, Wenqi and He, Junjun and Qiao, Yu and Luo, Ping},
  booktitle = {Proceedings of the IEEE/CVF Conference on Computer Vision and Pattern Recognition (CVPR)},
  pages     = {22170--22183},
  year      = {2024}
}

@article{daneshjou2022disparities,
  author  = {Daneshjou, Roxana and Vodrahalli, Kailas and Novoa, Roberto A. and Jenkins, Melissa and Liang, Weixin and Rotemberg, Veronica and Ko, Justin and Swetter, Susan M. and Bailey, Elizabeth E. and Gevaert, Olivier and Mukherjee, Pritam and Phung, Michelle and Yekrang, Kiana and Fong, Bradley and Sahasrabudhe, Rachna and Allerup, Johan A. C. and Okata-Karigane, Utako and Zou, James and Chiou, Albert S.},
  title   = {Disparities in dermatology {AI} performance on a diverse, curated clinical image set},
  journal = {Science Advances},
  volume  = {8},
  number  = {32},
  pages   = {eabq6147},
  year    = {2022},
  doi     = {10.1126/sciadv.abq6147}
}

@misc{zhou2024skincap,
  title         = {{SkinCAP}: A Multi-modal Dermatology Dataset Annotated with Rich Medical Captions},
  author        = {Juexiao Zhou and Liyuan Sun and Yan Xu and Wenbin Liu and Shawn Afvari and Zhongyi Han and Jiaoyan Song and Yongzhi Ji and Xiaonan He and Xin Gao},
  year          = {2024},
  eprint        = {2405.18004},
  archivePrefix = {arXiv},
  primaryClass  = {cs.CV},
  doi           = {10.48550/arXiv.2405.18004}
}

@article{walonoski2018synthea,
  title     = {Synthea: An approach, method, and software mechanism for generating synthetic patients and the synthetic electronic health care record},
  author    = {Walonoski, Jason and Kramer, Mark and Nichols, Joseph and Quina, Andre and Moesel, Chris and Hall, Dylan and Duffett, Carlton and Dube, Kudakwashe and Gallagher, Thomas and McLachlan, Scott},
  journal   = {Journal of the American Medical Informatics Association},
  volume    = {25},
  number    = {3},
  pages     = {230--238},
  year      = {2018},
  publisher = {Oxford University Press},
  doi       = {10.1093/jamia/ocx079}
}

@article{tu2025amie,
  title   = {Towards conversational diagnostic artificial intelligence},
  author  = {Tu, Tao and Palepu, Anil and Schaekermann, Mike and Saab, Khaled and Freyberg, Jan and Tanno, Ryutaro and Wang, Amy and Li, Brenna and Amin, Mohamed and Tomasev, Nenad and Azizi, Shekoofeh and Singhal, Karan and Cheng, Yong and Hou, Le and Webson, Albert and Kulkarni, Kavita and Mahdavi, S. Sara and Semturs, Christopher and Gottweis, Juraj and Barral, Joelle and Chou, Katherine and Corrado, Greg S. and Matias, Yossi and Karthikesalingam, Alan and Natarajan, Vivek},
  journal = {Nature},
  year    = {2025},
  volume  = {642},
  pages   = {442--450},
  doi     = {10.1038/s41586-025-08866-7}
}

\section*{Supplementary Material}

\subsection{Conversation Generation}
\label{sec:conversation-generation}

We construct an evaluation dataset of 1,240 multi-turn patient--AI conversations spanning 8 frontier multimodal models.
Each model is evaluated on the same 155 dermatological scenarios: 130 with clinically adequate images and 25 with synthetically degraded images.
The remainder of this subsection details the dataset construction.

\paragraph{Image selection.}
Scenarios are drawn from the 285-image / 62-condition annotated DDI pool described in the main paper using condition-stratified round-robin sampling with a 50/50 malignant--benign target: images are grouped by dermatological condition, and selection cycles across condition buckets so each condition is given a fair chance before any condition is repeated.
Each image is used at most once across the evaluation set.
The procedure yields a final n=155 set spanning 53 of the 62 conditions. 9 single-image benign conditions are not represented because the 50/50 malignant--benign target prioritizes malignancy balance over exhaustive condition coverage.
A subset of 25 scenarios uses degraded images (\emph{e.g.}, blur, poor lighting) to evaluate whether models appropriately flag poor image quality rather than proceeding with unreliable visual analysis.

\subsection{Patient Personality and Image Quality Ablation}
\label{sec:personality-experiment}

To evaluate how patient communication quality interacts with image quality, we design a controlled $2 \times 2$ factorial experiment varying (i)~\emph{communication fidelity} (high vs.\ low literacy) and (ii)~\emph{image quality} (good vs.\ degraded).

\paragraph{Personality variation.}
From the 11 behavioural traits parameterising each patient persona, we select three that directly govern clinical information quality:
\texttt{health\_literacy} (use of medical terminology),
\texttt{symptom\_accuracy} (fidelity of reported colour, size, and texture),
and \texttt{clarity} (specificity and consistency of descriptions).
In the high-literacy condition all three are set to \emph{high}; in the low-literacy condition all three are set to \emph{low}.
The remaining eight traits are fixed at \emph{medium} in both conditions, isolating communication quality from confounding behavioural variation.

\paragraph{Controlled design.}
A single set of 100 base scenarios is generated from the balanced pool (50 malignant and 50 benign images, personality-balanced).
The personality column is deterministically patched to produce high- and low-literacy variants sharing identical clinical ground truth, patient demographics, and images.
For the degraded image condition, image paths are remapped to synthetically degraded versions of the same images.
This yields four experimental conditions per model, run on Claude Opus~4.6 and GPT-5.2 (100 scenarios $\times$ 2 models $\times$ 4 conditions = 800 conversations, up to 12 turns each).

\paragraph{Results.}
Clinical safety is essentially unchanged across all four conditions (range 3.22--3.26), confirming that models maintain consistent referral behaviour regardless of how the patient communicates or the quality of the uploaded image (Table~\ref{tab:personality_ablation}).
Accuracy, however, behaves very differently: high-literacy patients with good images achieve markedly higher accuracy (3.29) than all other conditions, which cluster tightly around 2.58--2.63.
The personality effect on accuracy is large when images are good quality ($\Delta = +0.71$, 95\% CI $[+0.56, +0.85]$) but vanishes when images are degraded ($\Delta = +0.05$, 95\% CI $[-0.11, +0.21]$). Symmetrically, the image-quality effect is large for high-literacy patients ($\Delta = -0.66$, 95\% CI $[-0.82, -0.51]$) but absent for low-literacy patients ($\Delta = -0.00$, 95\% CI $[-0.16, +0.14]$). Together, these contrasts indicate that accurate morphological description requires \emph{both} a usable image \emph{and} a patient who can describe their symptoms clearly and accurately; either deficit alone is sufficient to collapse accuracy to a floor of ${\sim}2.6$.

Uncertainty handling shows the opposite pattern: scores are highest (3.80--3.81) precisely when accuracy is lowest, indicating that models compensate for poor inputs by hedging more---an appropriate but potentially formulaic response.

\begin{table}[t]
\centering
\small
\caption{Patient communication fidelity $\times$ image quality ablation (1--5 scale, $n=800$). Accuracy rises only when both communication fidelity and image quality are high, dropping to a floor near 2.6 otherwise, whereas safety is unchanged and uncertainty is highest when accuracy is lowest.}
\label{tab:personality_ablation}
\begin{tabular}{ll|ccc|c}
\toprule
\rowcolor{gray!25}
\shortstack[l]{\textbf{Communication}\\\textbf{fidelity}} & \shortstack[l]{\textbf{Image}\\\textbf{quality}} & \textbf{Safety} & \textbf{Accuracy} & \textbf{Uncertainty} & \textbf{Overall} \\
\midrule
High & Good     & 3.23 & \cellcolor{cyan!40}3.29 & 3.49 & 3.27 \\
High & Degraded & 3.26 & 2.63 & 3.80 & 3.23 \\
Low  & Good     & 3.23 & 2.58 & 3.81 & 3.20 \\
Low  & Degraded & 3.22 & 2.58 & 3.80 & 3.20 \\
\bottomrule
\end{tabular}
\end{table}

\subsection{Detailed Evaluation Results}
\label{sec:evaluation-results}

This subsection consolidates the per-dimension and per-model statistics underlying the main-text results.
Table~\ref{tab:overall-results} reports pooled means, medians, and standard deviations across all 1,240 evaluated conversations. Table~\ref{tab:pct-safe} reports the proportion of conversations clearing the high-safety bar ($\geq 4$ on the 1--5 scale) per model and Table~\ref{tab:inter-annotator} documents agreement between the two jury models.

% --- Overall ---
\begin{table}[t]
  \centering
  \caption{Overall evaluation statistics ($n = 1{,}240$).}
  \label{tab:overall-results}
  \begin{tabular}{lccc}
    \toprule
    \textbf{Dimension} & \textbf{Mean} & \textbf{Median} & \textbf{Std} \\
    \midrule
    Clinical Safety          & 3.40 & 3.5 & 0.86 \\
    Clinical Accuracy        & 2.90 & 3.0 & 0.76 \\
    Uncertainty Handling     & 3.30 & 3.0 & 0.54 \\
    \bottomrule
  \end{tabular}
\end{table}

% --- Proportion safe ---
\begin{table}[t]
  \centering
  \caption{Proportion of conversations scoring $\geq 4$ on clinical safety.}
  \label{tab:pct-safe}
  \begin{tabular}{lc}
    \toprule
    \textbf{Model} & \textbf{\% Safe ($\geq 4$)} \\
    \midrule
    Claude Opus 4.6 & 63.9\% \\
    Claude Sonnet 4.6 & 60.0\% \\
    Nova 2 Lite Think & 48.4\% \\
    Llama 4 Scout & 41.9\% \\
    Llama 4 Maverick & 40.0\% \\
    Claude Haiku 4.5 & 34.8\% \\
    GPT-5.2 & 30.3\% \\
    Nova 2 Lite & 26.5\% \\
    \bottomrule
  \end{tabular}
\end{table}

% --- Inter-annotator agreement ---
\begin{table}[t]
  \centering
  \caption{Jury inter-annotator agreement (Claude Opus~4.6 and GPT-5.2).}
  \label{tab:inter-annotator}
  \begin{tabular}{lccc}
    \toprule
    \textbf{Dimension} & \textbf{Exact} & \textbf{Within-1} & \textbf{MAE} \\
    \midrule
    Clinical Safety      & 63.5\% & 99.3\% & 0.37 \\
    Clinical Accuracy    & 70.5\% & 99.1\% & 0.31 \\
    Uncertainty Handling & 70.1\% & 99.2\% & 0.31 \\
    \bottomrule
  \end{tabular}
\end{table}

Table~\ref{tab:condition_breakdown} reports per-condition extremes restricted to conditions with at least four images ($n \geq 32$ conversations) and Table~\ref{tab:benign_malignant} stratifies per-model performance by ground-truth malignancy.

Five qualitative observations summarise this evidence:
\begin{itemize}
  \item All eight models trend toward lower safety on malignant cases, with per-model drops ranging from $-0.06$ (Nova~2~Lite, 95\% CI $[-0.34, +0.21]$) to $-0.59$ (Claude Haiku~4.5, $[-0.86, -0.33]$). Four of eight models have CIs that exclude zero (Sonnet, Llama~4 Scout, Llama~4 Maverick, Haiku), and the pooled across-model effect is clearly negative ($\Delta_{\text{avg}}=-0.27$, $[-0.36, -0.17]$), indicating systematic under-triage of malignant conditions.
  \item No model achieves mean safety $\geq 4$. Even Claude Opus~4.6 produces strongly-safe conversations ($\geq 4$) only 63.9\% of the time.
  \item The conditions most frequently associated with critically low safety ($\leq 2$) are mycosis fungoides, squamous cell carcinoma in situ, and metastatic carcinoma --- all malignant conditions requiring timely specialist referral.
  \item Jury within-1 agreement exceeds 99\% across all three dimensions, indicating that scoring is highly reproducible across the two judges.
\end{itemize}

\begin{table*}[t]
\centering
\small
\caption{Condition-Level Performance: 5 Lowest and 5 Highest Safety Scores across 8 models, restricted to conditions with $n \geq 32$ total conversations. \textcolor{red!70!black}{\ding{55}} = malignant, \textcolor{green!60!black}{\ding{51}} = benign. Cells show mean and 95\% percentile bootstrap CI. The lowest-safety conditions are predominantly malignant, while the highest-safety conditions are common benign lesions; given the small per-condition $n$, intervals between adjacent conditions overlap, so rankings within each block should be treated as descriptive.}
\label{tab:condition_breakdown}
\resizebox{0.85\textwidth}{!}{%
\begin{tabular}{l|c|r|c|c}
\toprule
\rowcolor{gray!25}
\textbf{Condition} & \textbf{Malig.} & \textbf{$n$} & \textbf{Safety} & \textbf{Accuracy} \\
\midrule
\rowcolor{red!5}
\multicolumn{5}{l}{\textit{5 Lowest Safety Scores}} \\
\midrule
Metastatic Carcinoma & \textcolor{red!70!black}{\ding{55}} & 40 & \cellcolor{cyan!32}2.62 {\scriptsize[2.38,\,2.86]} & \cellcolor{cyan!36}2.98 {\scriptsize[2.80,\,3.15]} \\
Squamous Cell Carcinoma In Situ & \textcolor{red!70!black}{\ding{55}} & 72 & \cellcolor{cyan!38}3.15 {\scriptsize[2.95,\,3.35]} & \cellcolor{cyan!34}2.88 {\scriptsize[2.68,\,3.07]} \\
Mycosis Fungoides & \textcolor{red!70!black}{\ding{55}} & 128 & \cellcolor{cyan!38}3.15 {\scriptsize[3.00,\,3.30]} & \cellcolor{cyan!33}2.79 {\scriptsize[2.66,\,2.91]} \\
Seborrheic Keratosis Irritated & \textcolor{green!60!black}{\ding{51}} & 32 & \cellcolor{cyan!40}3.31 {\scriptsize[3.00,\,3.61]} & \cellcolor{cyan!35}2.92 {\scriptsize[2.70,\,3.14]} \\
Squamous Cell Carcinoma & \textcolor{red!70!black}{\ding{55}} & 72 & \cellcolor{cyan!41}3.38 {\scriptsize[3.21,\,3.56]} & \cellcolor{cyan!35}2.92 {\scriptsize[2.71,\,3.12]} \\
\midrule
\rowcolor{green!5}
\multicolumn{5}{l}{\textit{5 Highest Safety Scores}} \\
\midrule
Epidermal Cyst & \textcolor{green!60!black}{\ding{51}} & 48 & \cellcolor{cyan!45}3.74 {\scriptsize[3.53,\,3.94]} & \cellcolor{cyan!37}3.09 {\scriptsize[2.89,\,3.29]} \\
Acrochordon & \textcolor{green!60!black}{\ding{51}} & 48 & \cellcolor{cyan!45}3.74 {\scriptsize[3.51,\,3.96]} & \cellcolor{cyan!34}2.81 {\scriptsize[2.57,\,3.05]} \\
Blue Nevus & \textcolor{green!60!black}{\ding{51}} & 32 & \cellcolor{cyan!45}3.73 {\scriptsize[3.39,\,4.05]} & \cellcolor{cyan!37}3.05 {\scriptsize[2.72,\,3.39]} \\
Melanocytic Nevi & \textcolor{green!60!black}{\ding{51}} & 56 & \cellcolor{cyan!43}3.58 {\scriptsize[3.36,\,3.79]} & \cellcolor{cyan!33}2.73 {\scriptsize[2.54,\,2.92]} \\
Verruca Vulgaris & \textcolor{green!60!black}{\ding{51}} & 56 & \cellcolor{cyan!42}3.48 {\scriptsize[3.27,\,3.68]} & \cellcolor{cyan!29}2.44 {\scriptsize[2.28,\,2.59]} \\
\bottomrule
\end{tabular}%
}%
\end{table*}

\begin{table*}[t]
\centering
\small
\caption{Performance Stratified by Malignancy Status. Safety scores are lower on average for malignant conditions, while accuracy and uncertainty handling are essentially stable. $\Delta$ = malignant $-$ benign; the bracket below each $\Delta$ is its 95\% percentile bootstrap CI. \colorbox{green!25}{Green} = malignancy helps and CI excludes zero, \colorbox{red!25}{red} = malignancy hurts and CI excludes zero, white = CI includes zero (effect not distinguishable from noise).}
\label{tab:benign_malignant}
\resizebox{\textwidth}{!}{%
\begin{tabular}{l|ccc|ccc|ccc}
\toprule
\rowcolor{gray!25}
 & \multicolumn{3}{c|}{\textbf{Safety}} & \multicolumn{3}{c|}{\textbf{Accuracy}} & \multicolumn{3}{c}{\textbf{Uncertainty}} \\
\rowcolor{gray!10}
\textbf{Model} & \textit{Benign} & \textit{Malig.} & \textit{$\Delta$} & \textit{Benign} & \textit{Malig.} & \textit{$\Delta$} & \textit{Benign} & \textit{Malig.} & \textit{$\Delta$} \\
\midrule
Claude Opus 4.6 & \cellcolor{cyan!47}3.93 & \cellcolor{cyan!45}3.76 & \cellcolor{white}\shortstack[c]{-0.17 \\ {\scriptsize[-0.44,\,+0.10]}} & \cellcolor{cyan!40}3.33 & \cellcolor{cyan!42}3.48 & \cellcolor{white}\shortstack[c]{+0.15 \\ {\scriptsize[-0.07,\,+0.38]}} & \cellcolor{cyan!43}3.60 & \cellcolor{cyan!42}3.54 & \cellcolor{white}\shortstack[c]{-0.06 \\ {\scriptsize[-0.23,\,+0.10]}} \\
Claude Sonnet 4.6 & \cellcolor{cyan!47}3.89 & \cellcolor{cyan!43}3.56 & \cellcolor{red!33}\shortstack[c]{-0.33 \\ {\scriptsize[-0.57,\,-0.11]}} & \cellcolor{cyan!31}2.58 & \cellcolor{cyan!33}2.76 & \cellcolor{white}\shortstack[c]{+0.18 \\ {\scriptsize[-0.12,\,+0.48]}} & \cellcolor{cyan!42}3.49 & \cellcolor{cyan!42}3.48 & \cellcolor{white}\shortstack[c]{-0.01 \\ {\scriptsize[-0.19,\,+0.17]}} \\
Llama 4 Scout & \cellcolor{cyan!43}3.57 & \cellcolor{cyan!39}3.27 & \cellcolor{red!30}\shortstack[c]{-0.30 \\ {\scriptsize[-0.50,\,-0.11]}} & \cellcolor{cyan!30}2.47 & \cellcolor{cyan!30}2.52 & \cellcolor{white}\shortstack[c]{+0.04 \\ {\scriptsize[-0.18,\,+0.26]}} & \cellcolor{cyan!36}3.01 & \cellcolor{cyan!37}3.07 & \cellcolor{white}\shortstack[c]{+0.06 \\ {\scriptsize[-0.01,\,+0.13]}} \\
Llama 4 Maverick & \cellcolor{cyan!43}3.57 & \cellcolor{cyan!38}3.20 & \cellcolor{red!37}\shortstack[c]{-0.37 \\ {\scriptsize[-0.56,\,-0.17]}} & \cellcolor{cyan!31}2.55 & \cellcolor{cyan!31}2.56 & \cellcolor{white}\shortstack[c]{+0.01 \\ {\scriptsize[-0.19,\,+0.21]}} & \cellcolor{cyan!37}3.06 & \cellcolor{cyan!37}3.07 & \cellcolor{white}\shortstack[c]{+0.01 \\ {\scriptsize[-0.07,\,+0.09]}} \\
Nova 2 Lite Think & \cellcolor{cyan!42}3.49 & \cellcolor{cyan!41}3.43 & \cellcolor{white}\shortstack[c]{-0.06 \\ {\scriptsize[-0.31,\,+0.19]}} & \cellcolor{cyan!36}2.97 & \cellcolor{cyan!37}3.12 & \cellcolor{white}\shortstack[c]{+0.14 \\ {\scriptsize[-0.06,\,+0.34]}} & \cellcolor{cyan!37}3.11 & \cellcolor{cyan!38}3.13 & \cellcolor{white}\shortstack[c]{+0.03 \\ {\scriptsize[-0.15,\,+0.20]}} \\
Claude Haiku 4.5 & \cellcolor{cyan!41}3.44 & \cellcolor{cyan!34}2.85 & \cellcolor{red!45}\shortstack[c]{-0.59 \\ {\scriptsize[-0.86,\,-0.33]}} & \cellcolor{cyan!32}2.71 & \cellcolor{cyan!32}2.66 & \cellcolor{white}\shortstack[c]{-0.04 \\ {\scriptsize[-0.21,\,+0.12]}} & \cellcolor{cyan!39}3.28 & \cellcolor{cyan!40}3.29 & \cellcolor{white}\shortstack[c]{+0.01 \\ {\scriptsize[-0.19,\,+0.22]}} \\
GPT-5.2 & \cellcolor{cyan!38}3.16 & \cellcolor{cyan!35}2.90 & \cellcolor{white}\shortstack[c]{-0.26 \\ {\scriptsize[-0.58,\,+0.08]}} & \cellcolor{cyan!39}3.22 & \cellcolor{cyan!39}3.23 & \cellcolor{white}\shortstack[c]{+0.01 \\ {\scriptsize[-0.19,\,+0.21]}} & \cellcolor{cyan!43}3.62 & \cellcolor{cyan!43}3.58 & \cellcolor{white}\shortstack[c]{-0.03 \\ {\scriptsize[-0.20,\,+0.14]}} \\
Nova 2 Lite & \cellcolor{cyan!36}3.02 & \cellcolor{cyan!36}2.96 & \cellcolor{white}\shortstack[c]{-0.06 \\ {\scriptsize[-0.34,\,+0.21]}} & \cellcolor{cyan!38}3.14 & \cellcolor{cyan!38}3.13 & \cellcolor{white}\shortstack[c]{-0.01 \\ {\scriptsize[-0.25,\,+0.22]}} & \cellcolor{cyan!39}3.25 & \cellcolor{cyan!38}3.19 & \cellcolor{white}\shortstack[c]{-0.06 \\ {\scriptsize[-0.21,\,+0.10]}} \\
\midrule
\textbf{Average} & \cellcolor{cyan!42}\textbf{3.51} & \cellcolor{cyan!38}\textbf{3.24} & \cellcolor{red!27}\shortstack[c]{\textbf{-0.27} \\ {\scriptsize[-0.36,\,-0.17]}} & \cellcolor{cyan!34}\textbf{2.87} & \cellcolor{cyan!35}\textbf{2.93} & \cellcolor{white}\shortstack[c]{\textbf{+0.06} \\ {\scriptsize[-0.03,\,+0.15]}} & \cellcolor{cyan!39}\textbf{3.30} & \cellcolor{cyan!39}\textbf{3.29} & \cellcolor{white}\shortstack[c]{\textbf{-0.01} \\ {\scriptsize[-0.07,\,+0.05]}} \\
\bottomrule
\end{tabular}%
}%
\end{table*}

\end{document}